\documentclass[10pt,twocolumn,letterpaper]{article}

\usepackage{cvpr}              
\usepackage{graphicx}
\usepackage{amsthm,amsmath,amssymb}
\usepackage{mathrsfs}
\usepackage{multirow}
\usepackage{mathtools}
\usepackage{bbding}

\definecolor{cvprblue}{rgb}{0.21,0.49,0.74}
\usepackage[pagebackref,breaklinks,colorlinks,allcolors=cvprblue]{hyperref}

\def\paperID{1847} 
\def\confName{CVPR}
\def\confYear{2025}

\title{GaussianIP: Identity-Preserving Realistic 3D Human Generation via Human-Centric Diffusion Prior}

\author{
  Zichen Tang\textsuperscript{\rm 1} \and 
  Yuan Yao \and 
  Miaomiao Cui \and 
  Liefeng Bo \and 
  Hongyu Yang\textsuperscript{\rm 1,2}\thanks{Corresponding author.} \and \vspace{-3mm} \\
  \textsuperscript{\rm 1}School of Artificial Intelligence, Beihang University, Beijing, China\\
  \textsuperscript{\rm 2}Shanghai Artificial Intelligence Laboratory, Shanghai, China\\
  {\tt\small \{zctang, hongyuyang\}@buaa.edu.cn}
}

\begin{document}
\maketitle
\begin{abstract}
Text-guided 3D human generation has advanced with the development of efficient 3D representations and 2D-lifting methods like Score Distillation Sampling (SDS). However, current methods suffer from prolonged training times and often produce results that lack fine facial and garment details. In this paper, we propose GaussianIP, an effective two-stage framework for generating identity-preserving realistic 3D humans from text and image prompts. Our core insight is to leverage human-centric knowledge to facilitate the generation process. In stage 1, we propose a novel Adaptive Human Distillation Sampling (AHDS) method to rapidly generate a 3D human that maintains high identity consistency with the image prompt and achieves a realistic appearance. Compared to traditional SDS methods, AHDS better aligns with the human-centric generation process, enhancing visual quality with notably fewer training steps. To further improve the visual quality of the face and clothes regions, we design a View-Consistent Refinement (VCR) strategy in stage 2. Specifically, it produces detail-enhanced results of the multi-view images from stage 1 iteratively, ensuring the 3D texture consistency across views via mutual attention and distance-guided attention fusion. Then a polished version of the 3D human can be achieved by directly perform reconstruction with the refined images. Extensive experiments demonstrate that GaussianIP outperforms existing methods in both visual quality and training efficiency, particularly in generating identity-preserving results. Our code is available at: \href{https://github.com/silence-tang/GaussianIP}{https://github.com/silence-tang/GaussianIP}.
\end{abstract}    
\section{Introduction}
\label{sec:intro}

Creating high-quality 3D human avatars based on user inputs is essential for a variety of applications, including Virtual Try-On~\cite{virtualtryon1, virtualtryon2, virtualtryon3, virtualtryon4} and immersive telepresence~\cite{telepresence1, telepresence2, telepresence3}. It also plays an pivotal role in emerging technologies like AR and VR. Text-guided 3D human generation is a task which involves synthesizing a character's geometry and appearance from text prompts. Score Distillation Sampling (SDS) proposed in DreamFusion~\cite{poole2022dreamfusion} paves the way for generating 3D objects by distilling from 2D diffusion priors~\cite{ho2020denoising, rombach2022high}, simplifying the process of creating 3D objects. It also inspires researchers to design various pipelines tailored for creating 3D humans.

A common paradigm is to learn a SDS-guided conditional Neural Radiance Field (NeRF)~\cite{mildenhall2021nerf} by integrating parametric human body models like SMPL~\cite{smpl} or imGHUM~\cite{alldieck2021imghum} into the framework to model body-related geometry and appearance~\cite{cao2024dreamavatar, kolotouros2024dreamhuman, huang2024dreamwaltz, zhang2024avatarverse}. However, NeRF-based methods suffer from slow rendering speed and fall short in delivering high-resolution results.

Recently, as 3D Gaussian Splatting (3DGS)~\cite{kerbl20233d} unlocks new possibilities for high-fidelity 3D reconstruction with real-time rendering, a few works~\cite{liu2024humangaussian, yuan2024gavatar} introduce 3DGS to achieve efficient creation of 3D avatars. Despite achieving promising results, all the NeRF-based or 3DGS-based methods rely solely on text prompts, limiting their diversity in generation and real-world applicability, such as creating identity-preserving 3D avatars from user portraits.

In contrast to these 3D-based methods, recent 2D generative diffusion models~\cite{ho2020denoising, rombach2022high} have exhibited pronounced superiority in terms of visual fidelity. Built upon these powerful generative base models, several works concentrate on human-centric generation tasks, e.g. Virtual Try-On (VTON)~\cite{morelli2023ladi, zhu2023tryondiffusion, kim2024stableviton, choi2024improving, ning2024picture, sun2024outfitanyone} and identity-driven photo personalization~\cite{ye2023ip, wang2024instantid, cui2024idadapter, li2024photomaker}. Although~\cite{chen2024gaussianvton, wang2024gaussianeditor, wu2024gaussctrl} attempt to edit 3D human bodies with the aid of 2D human-centric diffusion models, the potential of leveraging these models to generate detailed identity-preserving 3D humans from multi-modal user inputs, remains largely untapped.

In this work, we propose a novel two-stage framework GaussianIP, which is capable of generating realistic identity-preserving 3D human avatars from both text and image prompts. Our key intuition is twofold: (a) We can develop an effective method to distill human-centric diffusion priors, enabling the generation of 3D avatars with high facial identity consistency to the input image and rich clothing details; (b) The powerful generative capability of diffusion models can be leveraged to further refine the results of the distillation process. In stage 1, we train the 3D human Gaussians with our proposed \textbf{Adaptive Human Distillation Sampling (AHDS)} guidance, consisting of Human Distillation Sampling (HDS) and an adaptive timestep scheduling strategy. The HDS is designed by decomposing the original score difference and incorporating identity conditions to ensure identity-preserving distillation. In timestep scheduling, we treat the entire HDS process as three consecutive phases, from coarse geometry to fine facial details, and assign different diffusion timestep sampling strategies to each phase to accelerate the generation process. Since the distilled results may exhibit subtle texture smoothing, we design an elaborative refinement mechanism in stage 2 to further improve the intricate details of the clothed human body. Instead of diffusing and denoising the multi-view images directly, which may jeopardizing 3D consistency, we propose a \textbf{View-Consistent Refinement (VCR)} mechanism. Specifically, four main views are denoised first and their intermediate attention features are stored to guide the denoising process of the $k$ key views through mutual attention. For other ordinary views, we apply relative distance guided attention fusion to ensure texture consistency with their neighboring key views. Afterwards, the refined images can be utilized to optimize the 3D human Gaussians directly in a reconstruction way, which is highly efficient. Our key contributions include:
\begin{itemize}
\item We propose a novel identity-preserving 3D human generation task, where the human avatar should align with the given text and image prompts while maintaining highly-realistic face and clothes appearance. We design GaussianIP, a two-stage framework based on 3D Gaussian Splatting and 2D human-centric prior to achieve this goal.
\item We propose an Adaptive Human Distillation Sampling (AHDS) strategy to efficiently generate high-quality 3D human models in stage 1. Compared to the traditional SDS strategy, our AHDS better supports the human-specific generation process, yielding results with enhanced visual quality while reducing training steps by approximately 30\%.
\item We introduce a View-Consistent Refinement (VCR) mechanism to enhance the visual details of multi-view images from stage 1. Leveraging the refined images, the 3D human model can be further improved in a short time.
\item Extensive experiments show that GaussianIP outperforms existing methods in both visual quality and training efficiency, particularly excelling at generating identity-preserving facial details and rich garment textures.
\end{itemize}

\section{Related Work}
\begin{figure*}[t]
\centering
\includegraphics[width=1.0\textwidth]{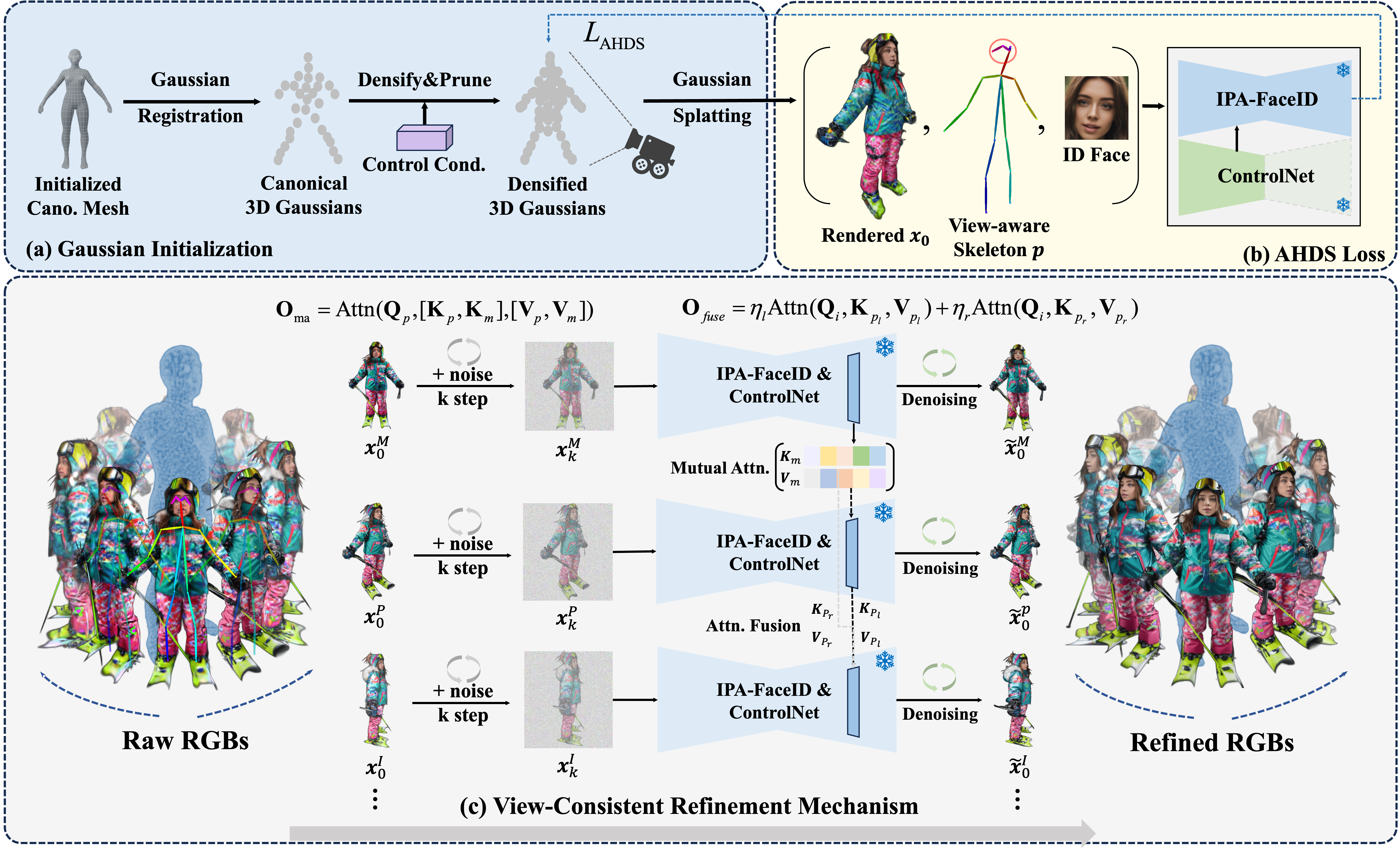}
\caption{Overview of the GaussianIP framework. We combine 3D Gaussian Splatting (3DGS) with a human-centric diffusion prior to realize high-fidelity 3D human avatar generation. (a) We initialize 3D human Gaussians by densely sample from a SMPL-X mesh. Afterward, (b) a human-centric diffusion model is combined with a pose-guide ControlNet to produce AHDS guidance. The AHDS guidance consists of an HDS guidance, which is proposed to achieve better identity-preserving generation, and an Adaptive Human-specific Timestep Scheduling strategy, which accelerates the HDS training. Furthermore, we propose (c) a View-Consistent Refinement Mechanism to further enhance the delicate texture of faces and garments. We guide the denoising of key views $\boldsymbol{x}_0^P$ with attention features from main views $\boldsymbol{x}_0^M$ through Mutual Attention. Next, we align the denoising of an intermediate view $\boldsymbol{x}_0^I$ with that of its neighbor key views via distance-guided attention fusion. Finally, the refined multi-view images are leveraged to optimize the current 3DGS.}
\label{fig:1}
\end{figure*}


\textbf{2D Human-centric Diffusion Models.} 2D generative diffusion models~\cite{ho2020denoising, rombach2022high, luo2023latent} have demonstrated remarkable performance in the realm of conditional image generation~\cite{ho2022classifier, gal2022image, zhang2023adding, ruiz2023dreambooth, kumari2023multi, shi2024instantbooth, xiao2024fastcomposer, mou2024t2i}, though they are typically designed for general object generation. Recently, researchers fine-tune these models on human-related datasets~\cite{xia2021tedigan, zheng2022general}, creating models that perform exceptionally well in generating content such as clothed human body. A prominent line of work is identity-preserving image customization~\cite{ye2023ip, wang2024instantid, liang2024caphuman, cui2024idadapter, he2024uniportrait, li2024photomaker}. These methods achieve single-ID personalization by leveraging global~\cite{radford2021learning} or local~\cite{deng2019arcface} facial features to condition the generation process on the identity of a single individual. Another noteworthy direction explores Virtual Try-On (VTON)~\cite{zhu2023tryondiffusion, choi2024improving, ning2024picture, li2024unihuman, sun2024outfitanyone}, aiming to generate an outfitted human wearing the given garment. These methods typically involve the integration of garment warping modules to generate deformed garments~\cite{morelli2023ladi, gou2023taming} or designing specialized UNet or injection blocks~\cite{xu2024ootdiffusion, kim2024stableviton} to align the garment features with the human body. However, lifting these 2D methods to 3D scenes, thereby expanding their applications in AR/VR, remains largely unexplored.

\textbf{Text-guided 3D General Object Generation.} Traditional methods~\cite{sanghi2022clip, jain2022zero, mohammad2022clip} rely on CLIP~\cite{radford2021learning} to optimize the 3D representations while decent results cannot be achieved due to the low expressivity of CLIP. Score Distillation Sampling (SDS) proposed in DreamFusion~\cite{poole2022dreamfusion} opens the door for high-quality 3D object generation by distilling from text-to-image diffusion models and motivates various concurrent works~\cite{wang2023score, lin2023magic3d, raj2023dreambooth3d, chen2023fantasia3d}. To improve the original SDS loss, several novel distillation techniques~\cite{huang2023dreamtime, yu2023text, katzir2023noise, lukoianov2024score, chen2024vividdreamer} have been introduced, such as Variational Score Distillation (VSD)~\cite{wang2024prolificdreamer}, Internal Score Matching (ISM)~\cite{liang2024luciddreamer}, Asynchronous Score Distillation (ASD)~\cite{ma2025scaledreamer}, etc., by reexamining the SDS process or analyzing the behavior of 2D diffusion models. Considering the time-consuming NeRF training process, several works~\cite{tang2023dreamgaussian, yi2024gaussiandreamer, melas20243d} adopt 3D Gaussian Splatting (3DGS)~\cite{kerbl20233d} as their 3D representation to achieve efficient training and rendering. 

\textbf{Text-guided 3D Human Generation.} AvatarCLIP~\cite{hong2022avatarclip} first realizes zero-shot text-driven 3D avatar generation by leveraging CLIP and NeuS~\cite{wang2021neus}. Inspired by SDS, DreamWaltz~\cite{huang2024dreamwaltz}, AvatarVerse~\cite{zhang2024avatarverse} and other concurrent works~\cite{kolotouros2024dreamhuman, cao2024dreamavatar} utilize the parametric SMPL model~\cite{smpl} and auxiliary networks like densepose ControlNet~\cite{zhang2023adding} to guide the learning process, improving the visual fidelity and 3D consistency of the generated 3D avatars. Apart from these NeRF-based methods, some works integrate other 3D representations into their frameworks to attain more fine-grained control of geometric shapes or textures. AvatarCraft~\cite{jiang2023avatarcraft}, TADA~\cite{liao2024tada} and X-Oscar~\cite{ma2024x} employ optimized SMPL-X mesh to represent 3D human bodies. AvatarStudio~\cite{mendiratta2023avatarstudio} and HumanNorm~\cite{huang2024humannorm} achieve enhanced geometric details by utilizing DMTet~\cite{shen2021deep} for 3D representation, combined with multi-stage or decoupled optimization strategies. More recently, HumanGaussian~\cite{liu2024humangaussian}, GAvatar~\cite{yuan2024gavatar} and other works~\cite{liu2024clothedreamer, gong2024laga} introduce 3DGS to create high-resolution realistic 3D avatars and real-time rendering. However, these methods fail to handle image prompts such as user portraits, which limits their real-world applicability.
\section{Method}

\textbf{Problem statement.} Given a text prompt describing the clothing and an user portrait providing facial identity, the model should swiftly generate a identity-preserving 3D human avatar with refined facial features and intricate clothing details. In the following sections, preliminaries will be present first. Next, we will introduce our Adaptive Human Distillation Sampling (AHDS) process. Following that, a simple yet effective View-Consistent Refinement mechanism (VCR) will be detailed.

\subsection{Preliminaries}

\noindent \textbf{3D Gaussian Splatting}~\cite{kerbl20233d} is an effective point-based representation consisting of a set of anisotropic Gaussians. Each 3D Gaussian is parameterized by its center position $\boldsymbol{\mu} \in \mathbb{R}^3$, covariance matrix $\boldsymbol{\Sigma} \in \mathbb{R}^7$, opacity $\alpha \in \mathbb{R}$ and color $\boldsymbol{c} \in \mathbb{R}^3$. By splatting 3D Gaussians onto 2D image planes, we can perform point-based rendering:
\begin{equation}\label{eq:1}
\begin{gathered}
G(\boldsymbol{p}, \boldsymbol{\mu}_{i}, \boldsymbol{\Sigma}_{i}) = \exp(- \frac{1}{2}(\boldsymbol{p} - \boldsymbol{\mu}_{i})^\intercal \boldsymbol{\Sigma}_{i}^{-1}(\boldsymbol{p} - \boldsymbol{\mu}_{i})), \\
\boldsymbol{c}(\boldsymbol{p}) = \sum\limits_{i\in \mathcal{N}} \boldsymbol{c}_{i}\alpha_{i}' \prod\limits_{j=1}^{i-1} (1 - \alpha_{j}'), \alpha_{i}' = \alpha_{i} G(\boldsymbol{p}, \boldsymbol{\mu}_{i}, \boldsymbol{\Sigma}_{i}).
\end{gathered}
\end{equation}

\noindent Here, $\boldsymbol{p}$ is the coordinate of the queried point. $\boldsymbol{\mu}_i$, $\boldsymbol{\Sigma}_i$, $\boldsymbol{c}_i$, $\alpha_i$, and $\alpha_i'$ denote the center, covariance, color, opacity, and density of the $i$-th Gaussian, respectively. $G(\boldsymbol{p}, \boldsymbol{\mu}_i, \boldsymbol{\Sigma}_i)$ represents the value of the $i$-th Gaussian at position $\boldsymbol{p}$. $\mathcal{N}$ is a sorted list of Gaussians in this tile.

\noindent \textbf{Score Distillation Sampling}~\cite{poole2022dreamfusion} leverages a powerful 2D text-to-image diffusion model, $\epsilon_\phi$, to guide the training of 3D scenes. By distilling gradients from the 2D diffusion model, SDS ensures the 3D scene's rendered images align with the input text prompt. Given a 3D representation parameterized by $\theta$, an image $\boldsymbol{x} = g(\theta)$ can be rendered with a differentiable renderer $g$. SDS calculates the gradients with respect to the 3D representation $\theta$ by:
\begin{equation}\label{eq:2}
    \nabla_\theta \mathcal{L}_{SDS}(\phi, \boldsymbol{x}) = \mathbb{E}_{t, \epsilon} \left[ w(t) \left( \epsilon_\phi(\boldsymbol{x}_t; y, t) - \boldsymbol{\epsilon} \right) \frac{\partial \boldsymbol{x}}{\partial \theta} \right],
\end{equation}

\noindent where $y$ is the text prompt, $t$ is the sampling timestep in a 2D diffusion model, $\boldsymbol{\epsilon} \sim \mathcal{N}(0, \mathbf{I})$ is the sampled noise and $\boldsymbol{x}_t$ is the noised image. $w(t)$ is a weighting function depending on the timestep $t$.

\begin{figure*}[t]
\centering
\includegraphics[width=1.0\textwidth]{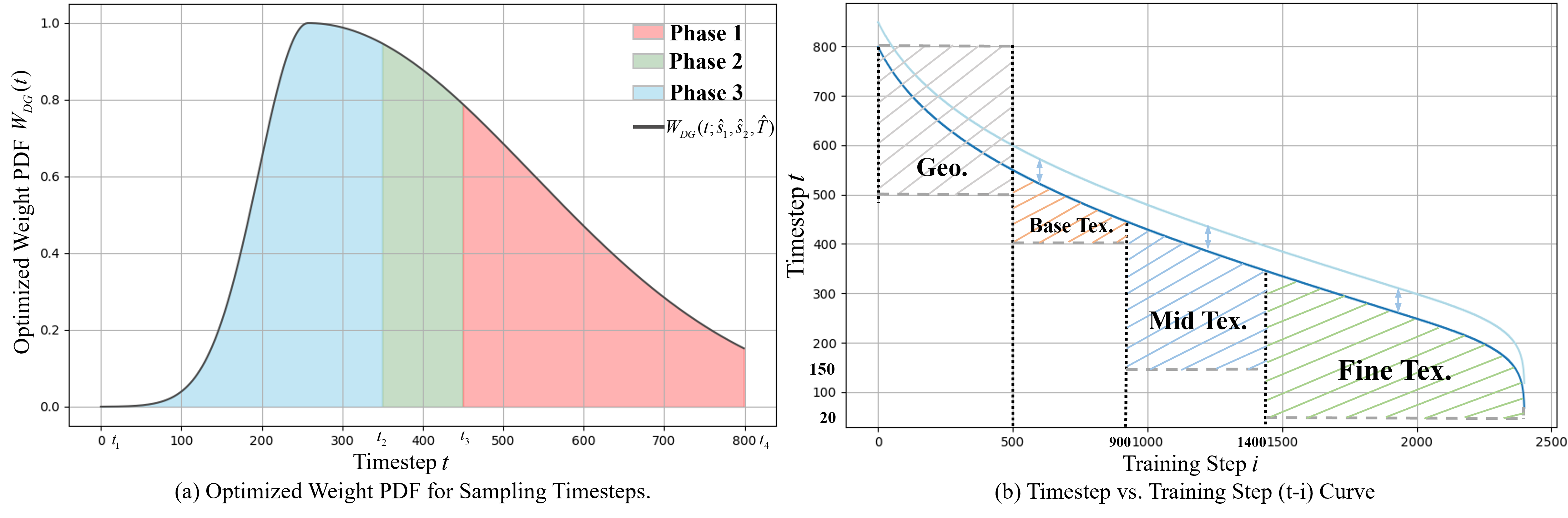}
\caption{Illustration of the optimized weight PDF for sampling timesteps and the corresponding timestep vs. training step (t-i) curve. a) Phase 1, 3 occupy the majority of the training steps, while Phase 2 occupies only a small portion, allowing a quick transition to the detailed texture learning in Phase 3. b) We sample the final timestep between the lower bound and $t_{\text{DG}}$ for each phase. Note that for the geometry phase ($i<500$), we sample between 500 and the maximum timestep to ensure a smooth start.}
\label{fig:2}
\end{figure*}

\subsection{Adaptive Human Distillation Sampling}

3D human generation task has distinct characteristics compared to 3D object generation tasks, which typically involve initializing 3D Gaussians~\cite{kerbl20233d} with Shape-e~\cite{jun2023shap} and optimizing the 3D object representation through distillation from 2D general-purpose diffusion models~\cite{ho2020denoising, rombach2022high}. For instance, the human parametric model~\cite{smplx} provides a reliable prior for human body geometry, which can facilitate the training process by starting with smaller diffusion timestep. By leveraging 2D human-centric diffusion models~\cite{ye2023ip, yan2023facestudio, wang2024instantid}, we can enhance the visual quality of generated 3D avatars and ensure identity-preserving generation through effective distillation techniques. In addition, inspired by the coarse to fine nature of 2D human generation process, we can design a more effective timestep sampling strategy tailored for human-specific generation tasks.

\noindent \textbf{Gaussian Initialization.} A proper initialization method which can provide a rough geometry is crucial for the success of training. Traditional methods~\cite{kerbl20233d, yi2024gaussiandreamer} rely on tools like Structure-from-Motion (SfM)~\cite{schonberger2016structure} and Shape-e~\cite{jun2023shap}, which may result in overly sparse points or inconsistent body structures. \cite{liu2024humangaussian} first proposes to initialize 3D Gaussians on a neutral SMPL-X~\cite{smplx} mesh. We adopt a similar way to prepare for the subsequent training stages.

\noindent \textbf{Distillation Sampling with Human-centric Prior.} Since we focus on 3D human generation, distilling from a general diffusion prior like~\cite{yuan2024gavatar, liu2024humangaussian, liu2024clothedreamer} may not be the most optimal approach. Instead, we combine a face-focused version of ~\cite{ye2023ip} $D_{ip}$ with a pose-conditioned ControlNet~\cite{zhang2023adding} $C_{pose}$ to form a human-specific diffusion prior $D_{ip}^p$. For each training step, we apply a view-dependent pose skeleton trimming strategy to ensure a pose condition map $\boldsymbol{p}$ which accurately represents the visibility of facial features (eyes, ears, etc.) from different views. For example, when the azimuth exceeds $60^{\circ}$, the left ear keypoint will be masked. Then the left and right eye will be sequentially masked as the azimuth increases. For the back view, only the ears remain visible. The trimmed $\boldsymbol{p}$ contains pose clue across views, which is essential for mitigating the ``Janus'' problem. \\
To fully leverage the given text and image condition when training with SDS loss, we propose a novel human distillation sampling (HDS) guidance by decomposing and redesigning the original score difference. Inspired by~\cite{katzir2023noise}, we can rewrite the common score difference~\cite{poole2022dreamfusion} as:
\begin{equation}\label{eq:3}  
    \epsilon_{\phi}^s(\boldsymbol{x}_t;y,t) - \boldsymbol{\epsilon} = \delta_{\text{rect}} + \delta_{\text{noise}} + \gamma\delta_{\text{cond}} - \boldsymbol{\epsilon},
\end{equation}
where $\delta_{\text{rect}} + \delta_{\text{noise}} = \epsilon_{\phi}(\boldsymbol{x}_t;y_\phi,t)$ and $\delta_{\text{cond}} = \epsilon_{\phi}(\boldsymbol{x}_t;y,t) - \epsilon_{\phi}(\boldsymbol{x}_t;y_\phi,t)$ denote the unconditional and conditional terms, respectively, with $\gamma$ as the classifier-free guidance (CFG)~\cite{ho2022classifier} coefficient and $y_\phi$ the null text. Here, $\delta_{\text{rect}}$ acts as a rectifier, guiding the rendered image $\boldsymbol{x} = g(\theta)$ towards a denser region within the manifold of real images, while $\delta_{\text{noise}}$ serves as a denoiser, pushing $\boldsymbol{x}$ towards a cleaner image. However, ~\cite{wang2024prolificdreamer} indicates that the residual $\delta_{\text{noise}}-\boldsymbol{\epsilon}$ is noisy and may cause blurry textures, so we define the identity-preserving score difference $\delta_{ip} = \epsilon_{ip}^s(\boldsymbol{x}_t;y,\boldsymbol{I}_{ip},t) - \boldsymbol{\epsilon}$ in our HDS as:
\begin{equation}\label{eq:4}
\begin{split}
     \delta_{ip} &= \delta_{\text{rect}}^{ip} + \gamma\delta_{\text{cond}}^{ip} \\
    &= \delta_{\text{rect}}^{ip} + \gamma(\epsilon_{ip}(\boldsymbol{x}_t;y,\boldsymbol{I}_{ip},t) - \epsilon_{ip}(\boldsymbol{x}_t;y_{\phi},\boldsymbol{I}_{\mu},t)),
\end{split}
\end{equation}
where $\epsilon_{ip}$ is the $\boldsymbol{\epsilon}$-prediction UNet used in the diffusion prior $D_{ip}^{pose}$ and $\boldsymbol{I}_\mu$ can be a face irrelevant to $\boldsymbol{I}_{ip}$. As regards the rectifying direction $\delta_{\text{rect}}$, it can be estimated as $\epsilon_{\phi}(\boldsymbol{x}_t;y_\phi,t)$ when the timestep $t$ is rather small, given the low noise level at these timesteps making the denoising direction negligible. We then use a repelling score to model $\delta_{rect}$ at other diffusion timesteps, so we get:
\begin{equation}\label{eq:5}
     \delta_{\text{rect}}^{ip} = 
\begin{cases}
\epsilon_{ip}(\boldsymbol{x}_t; y_\phi, \boldsymbol{I}_{\mu}, t)& \hspace{-0.8em} t < \tau \\
\epsilon_{ip}(\boldsymbol{x}_t; y_\phi, \boldsymbol{I}_{\mu}, t) - \epsilon_{ip}(\boldsymbol{x}_t; y_{\text{-}}, \boldsymbol{I}_\phi, t)& \hspace{-0.8em} t \geq \tau,
\end{cases}
\end{equation} 
where $y_{-}$ represents the negative prompt, $\boldsymbol{I}_\phi$ denotes an all-zero image, and $\tau$ is the timestep threshold. With our redesigned score difference, the generated avatar is more realistic without the issue of over-saturation and exhibits a strong alignment with the given identity image prompts.

\noindent \textbf{Adaptive Human-specific Timestep Scheduling.} Despite an identity-preserved 3D avatar with decent appearance can be generated with HDS, it still requires a long training time. Aims to achieve faster HDS optimization within fewer training steps, we propose an adaptive diffusion timestep scheduling strategy motivated by~\cite{huang2023dreamtime, lee2024dreamflow}, which results in a non-increasing timestep against training step ($t-i$) curve tailored for 3D human generation tasks. Inspired by the denoising process in 2D human generation, we may naturally divide the entire 3D HDS process into three synergistic phases: geometry and base textures (phase 1), middle-level textures (phase 2) and fine facial and garment details (phase 3). Each phase is assigned with a timestep range, $\text{range}_i = [t_i, t_{i+1}], i=1,2,3$. Intuitively, more training steps should be assigned to phase 1, 3 to model human geometry and intricate details. Phase 2 can be seen as a transitional stage, requiring slightly fewer steps. To derive a $t-i$ curve that accurately exhibits these trends, we start by optimizing a weighting PDF function of timesteps $t$ then map it to the final $t-i$ curve. Empirically, a dual-piecewise Gaussian function $W_{DG}(t;s_1, s_2, T)$ is employed to represent the probability distribution:
\begin{equation}\label{eq:6}
    W_{\text{DG}}(t) = 
    \begin{cases} 
    \frac{1}{\sqrt{2 \pi s_1^2}}\exp\left(-\frac{(t - T)^2}{2 s_1^2}\right)& t \leq T \\
    \frac{1}{\sqrt{2 \pi s_2^2}}\exp\left(-\frac{(t - T)^2}{2 s_2^2}\right)& t > T.
    \end{cases}
\end{equation}
The long-tail property may cause a swift timestep decrease at phase 1 to avoid over-large $t$ (SMPL-X provides a rough geometry so the training can start with smaller $t$) and suppress over-small $t$ at phase 3, which introduces high gradient variance. Given the training steps assigned to the three phases $n_i, i=1,2,3$ and the total step $N$, We can estimate the parameters of $W_{DG}$ by solving an optimization problem:
\begin{equation}\label{eq:7}
    \hat{s_1}, \hat{s_2}, \hat{T} = \underset{s_1, s_2, T}{\operatorname{argmin}} \sum_{k=1}^{3}\bigg(\sum_{t\in \text{range}_k}\!\!\!W_{\text{DG}}(t) - \frac{n_k}{N} \bigg)^2,
\end{equation}
with the results presented in Fig. \ref{fig:2} (a). This objective ensures that the accumulative probability of each range aligns with the desired training step proportion of each phase, facilitating the mapping between $W_{DG}$ and the $t-i$ curve. Finally, the corresponding timestep for each training step $i$ can be solved by:
\begin{equation}\label{eq:8}
  t_{\text{DG}}(i) = \underset{\tau}{\operatorname{argmin}} \left| \sum_{t=\tau}^{T} W_{DG}(t) - \frac{i}{N} \right| + \epsilon,
\end{equation}
where $\epsilon$ is a small tuning factor. To prevent texture over-saturation and ensure smooth transitions between stages, we set a lower bound for each phase and sample the final timestep within the range between this lower bound and $t_{\text{DG}}$ (shown in Fig. \ref{fig:2} (b)). This adaptive timestep scheduling function can be seamlessly integrated into our HDS process to form the AHDS guidance, ensuring a faster and smoother training process that produces detail-enhanced results.

\subsection{View-consistent Refinement Mechanism}\label{sec:3.3}
With the assistance of the AHDS guidance in stage 1, 3D avatars with rich garment details and better identity consistency can be generated efficiently. However, the distillation process itself may inevitably cause slight texture smoothing. To further enhance detail articulation based on the AHDS training results, a straightforward approach is to add noise to the rendered multi-view images, denoise them, and then optimize the 3DGS using the denoised images. The independent denoising process, however, may lead to texture inconsistencies across multi-view images, as there is no exchange of view information during processing. To address this issue, we design an elegant refinement strategy, which is shown in Fig. \ref{fig:1} (c). The entire refinement process consists of two sub-process, key view refinement and intermediate features propagation. We first guide the denoising of $k$ key views with the attention features of main views (front, back, left, right), ensuring a roughly consistent appearance among key and main views, then we propagate the refinement effects to other views via weighted attention fusion.

\textbf{Key Views Refinement.} For a specific key view $P$, we inject the attention keys $\boldsymbol{K}_m$ and values $\boldsymbol{V}_m$ of its nearest main view $M$ into the denoising process. Directly applying this injection may cause texture drift, as some unwanted features (from invisible areas) may be queried. To mitigate this, we inflate the self-attention keys and values for the main view, so the keys and values from the two views act as a mutual reference. Denote the attention operation as $\text{Attn}(\boldsymbol{Q}, \boldsymbol{K}, \boldsymbol{V}) = \text{Softmax}\left(\frac{\boldsymbol{Q}\boldsymbol{K}^\mathrm{T}}{\sqrt{d}}\right)\boldsymbol{V}$, where $d$ is the model dimension, we can denote the main view guided mutual attention as:
\begin{equation}\label{eq:9}
  \boldsymbol{O}_{ma} = \text{Attn}(\boldsymbol{Q}_p,[\boldsymbol{K}_p, \boldsymbol{K}_m], [\boldsymbol{V}_p, \boldsymbol{V}_m]),
\end{equation}
where $\boldsymbol{K}_p$, $\boldsymbol{V}_p$ denote the keys and values of view $P$, respectively, and $[\cdot, \cdot]$ is the concatenate operation. This mutual reference attention ensures that all the key views share an aligned appearance with the main views, laying a solid foundation for the following process.

\noindent \textbf{Intermediate Features Propagation.} Finally, to obtain a smooth transition of refinement effects between two adjacent key views (we may regard main views as key views here), we ``propagate'' the attention features of two key adjacent views to their intermediate views. Given such a view $I$, we first query its nearest neighbor key views $P_l$ and $P_r$ and calculate the relative distance based on their azimuths, that is: $\text{Dist}(I, P_l) = \frac{|\phi_{I} - \phi_{P_l}|}{|\phi_{P_r} - \phi_{P_l}|}$, where $\phi_V$ is the azimuth of view $V$. We can naturally regard $\eta_{l} = 1-Dist(I, P_l)$ as the influence of $P_l$ on $I$. To properly fuse the attention features $\boldsymbol{K}_{p_l}, \boldsymbol{V}_{p_l}$ of $P_l$ and $\boldsymbol{K}_{p_r}, \boldsymbol{V}_{p_r}$ of $P_r$ into the denoising process of $I$, we introduce a novel weighted attention fusion strategy guided by relative distances. Specifically, the attention output is combined with two parts, pure self-attention $\boldsymbol{O}_{\text{sa}} = \text{Attn}(\boldsymbol{Q}_i, \boldsymbol{K}_i, \boldsymbol{V}_i)$ and fused attention, which is obtained by $\boldsymbol{O}_{\text{fa}} = \eta_{l}\text{Attn}(\boldsymbol{Q}_i, \boldsymbol{K}_{P_l}, \boldsymbol{V}_{P_l}) + \eta_{r}\text{Attn}(\boldsymbol{Q}_i, \boldsymbol{K}_{P_r}, \boldsymbol{V}_{P_r})$. The output of the complete attention operation can then be derived as:
\begin{equation}\label{eq:10}
  \boldsymbol{O}_{\text{final}} = \lambda_{\text{self}}\boldsymbol{O}_{\text{sa}} + (1 - \lambda_{\text{self}})\boldsymbol{O}_{\text{fa}},
\end{equation}
where $\lambda_{\text{self}}\in[0,1]$ is a weight factor. Through this attention fusion process, the refined image of $I$ will have high texture consistency with the results of its neighboring views.

\noindent \textbf{3D Human Gaussian Optimization.} With the refined multi-view images which are aligned with each other in textures and semantics, we can optimize the 3D Human Gaussians from stage 1 by directly applying reconstruction losses. At each training step, we randomly select $b$ views from those used in the refinement stage and compute the reconstruction loss between the rendered images $\hat{\boldsymbol{X}}$ and the ground truth $\boldsymbol{X_{\text{gt}}}$:
\begin{equation}\label{eq:11}
  \mathcal{L}_{\text{recon}} = \lambda_{\text{L1}} \, L_1(\hat{\boldsymbol{X}}, \boldsymbol{X}_{\text{gt}}) + \lambda_{\text{lpips}} \, L_{\text{lpips}}(\hat{\boldsymbol{X}}, \boldsymbol{X}_{\text{gt}}),
\end{equation}
where $\lambda_{\text{L1}}$ and $\lambda_{\text{lpips}}$ are weight factors. To improve efficiency, we crop each rendered image to the rectangular region of the human body, downsample it by half, and then calculate the reconstruction loss against similarly processed ground truth images.
\section{Experiments}

\begin{figure*}[t]
\centering
\includegraphics[width=1.0\textwidth]{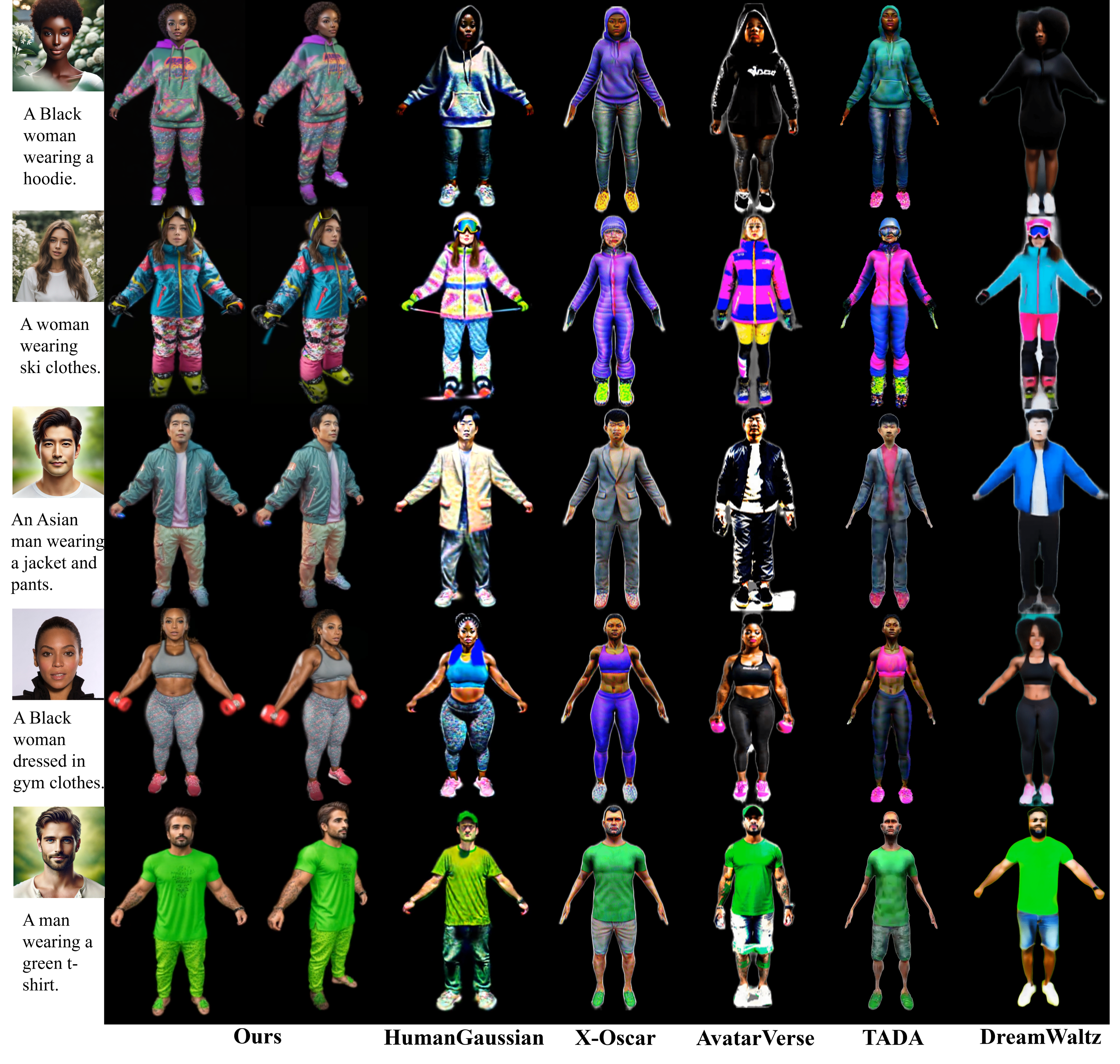}
\caption{Qualitative comparison results with SOTA text-guided 3D human generation models. Please zoom in for better observation. Note that the baselines cannot handle image prompts, so we compare their text-to-3D results instead. Due to space limitations, please refer to the supplementary materials for the video comparison results.}
\label{fig:3}
\end{figure*}

\begin{table*}[tb]
\centering
\begin{tabular}{l@{\hskip 4pt}c@{\hskip 4pt}c@{\hskip 4pt}c@{\hskip 4pt}c@{\hskip 4pt}c@{\hskip 4pt}c@{\hskip 4pt}c}
\hline
\textbf{Methods} & \textbf{Fac. Det.} & \textbf{Cloth. Tex.} & \textbf{Vis. Qual.} & \textbf{Text Align.} & \textbf{GPT Score} & \textbf{Training Time} & \textbf{ID Pres.} \\
\hline
DreamWaltz~\cite{huang2024dreamwaltz}      & 1.33 & 1.46 & 1.38 & 1.58 & 1.82 & 1.3h & \XSolidBrush \\
TADA~\cite{liao2024tada}              & 2.21 & 2.46 & 2.58 & 3.13 & 3.24 & 2h & \XSolidBrush \\
AvatarVerse~\cite{zhang2024avatarverse}    & 3.67 & 3.29 & 2.63 & 2.96 & 3.05 & 1h & \XSolidBrush \\
X-Oscar~\cite{ma2024x}        & 3.29 & 3.83 & 3.58 & 3.63 & 3.64 & 3h & \XSolidBrush \\
HumanGaussian~\cite{liu2024humangaussian} & 4.29 & 4.17 & 3.96 & 4.42 & 4.08 & 1.2h & \XSolidBrush \\
\textbf{Ours}                         & \textbf{4.71} & \textbf{4.50} & \textbf{4.17} & \textbf{4.62} & \textbf{4.52} & \textbf{40min} & \Checkmark \\
\hline
\end{tabular}
\caption{\textbf{Quantitative comparison results.} We conduct evaluations on generation quality from five aspects: (1) Facial Detail; (2) Clothing Texture Richness; (3) Overall Visual Quality; (4) Text Prompt Alignment; (5) ChatGPT Score. We also compare the training time and identity preserving ability here.}
\label{tab:1}
\end{table*}

\subsection{Implementation Details}

\textbf{AHDS Guidance and Stage 1 Training Setups.} We adopt IP-Adapter-FaceID-PlusV2~\cite{ye2023ip} as our human-centric diffusion prior. When solving the time scheduling $t-i$ curve, we set $t_1=20$, $t_2=350$, $t_3=450$ and $t_4=800$, respectively. The training steps assigned to the three phases are $n_1=900$, $n_2=500$ and $n_3=1000$, respectively, and the total step of phase 1 training is 2400. During the training process, we set the classifier-free guidance coefficient to $\gamma=7.5$ and the time threshold in Eq. \ref{eq:6} to $\tau=170$. The lower bounds for the three stages are 400, 150 and 20, respectively. The densification \& pruning operation is conducted from 200 to 1700 steps at an interval of 800 steps while the prune-only operation is applied once at step 1800. Stage 1 is trained with Adam optimizer, with the learning rates of each Gaussian attribute is scheduled following~\cite{liu2024humangaussian}. 

\noindent \textbf{Refinement and Stage 2 Training Setups.} The total denoising step is set to 8 during the refinement process and the number of key views is 4. When conducting intermediate view refinement, the weighted attention fusion factor $\lambda_{\text{self}}$ is set to 0.55. We use the refined images to optimize the 3D human Gaussians for 800 steps with a batch size $b=8$. The weight factors $\lambda_{\text{L1}}$ and $\lambda_{\text{lpips}}$ are set to 10 and 15, respectively. All experiments are conducted on a single NVIDIA V100 GPU.

\noindent \textbf{Baselines.} For fair comparison, we compare with recent SOTA text-to-3D human works instead of general text-to-3D methods, which include DreamWaltz~\cite{huang2024dreamwaltz}, TADA~\cite{liao2024tada}, AvatarVerse~\cite{zhang2024avatarverse}, X-Oscar~\cite{ma2024x} and HumanGaussian~\cite{liu2024humangaussian}.

\noindent \textbf{Criteria.} We conduct a user study to evaluate the 3D human humans generated by different methods. We randomly selected 20 prompts to generate 3D human models using each of the comparison methods. 24 participants were asked to evaluate the generated models based on 4 criteria: Facial Detail, Clothing Texture Richness, Overall Visual Quality, and Text Prompt Alignment. We also have ChatGPT-4o rate the images comprehensively on these aspects. Each criterion was rated on a scale from 1 to 5 (higher is better). Additionally, we compare the GPU memory cost and training time of each method to assess training efficiency.

\subsection{Text-guided 3D Human Generation}

\textbf{Qualitative Results.} Our method demonstrates clear advantages over~\cite{liao2024tada, kolotouros2024dreamhuman, cao2024dreamavatar, liu2024humangaussian} in two key aspects. As shown in Fig. \ref{fig:3}, it excels in facial detail, capturing sharper and more realistic facial features. For clothing texture richness, our method preserves intricate patterns and fabric details that others often miss. Additionally, our method generates results with high identity consistency to the given user portraits, which broadens its potential application scenarios.

\noindent \textbf{Quantitative Results.} As shown in Table \ref{tab:1}, our method achieves consistently higher scores across all five criteria, demonstrating superior fidelity and alignment with prompts. Additionally, We utilize an off-the-shelf commercial face verification algorithm (Face++) to evaluate the performance of our method with regard to identity preservation. All faces of the generated humans successfully match the corresponding user portraits with an average confidence level exceeding 83\%. Regarding training time, all baseline methods take over 1 hour, while our method completes training in just 40 minutes, demonstrating its superior efficiency. Moreover, our method requires less GPU memory (less than 24GB) than other methods.

\begin{figure}[t]
\centering
\includegraphics[width=1.0\columnwidth]{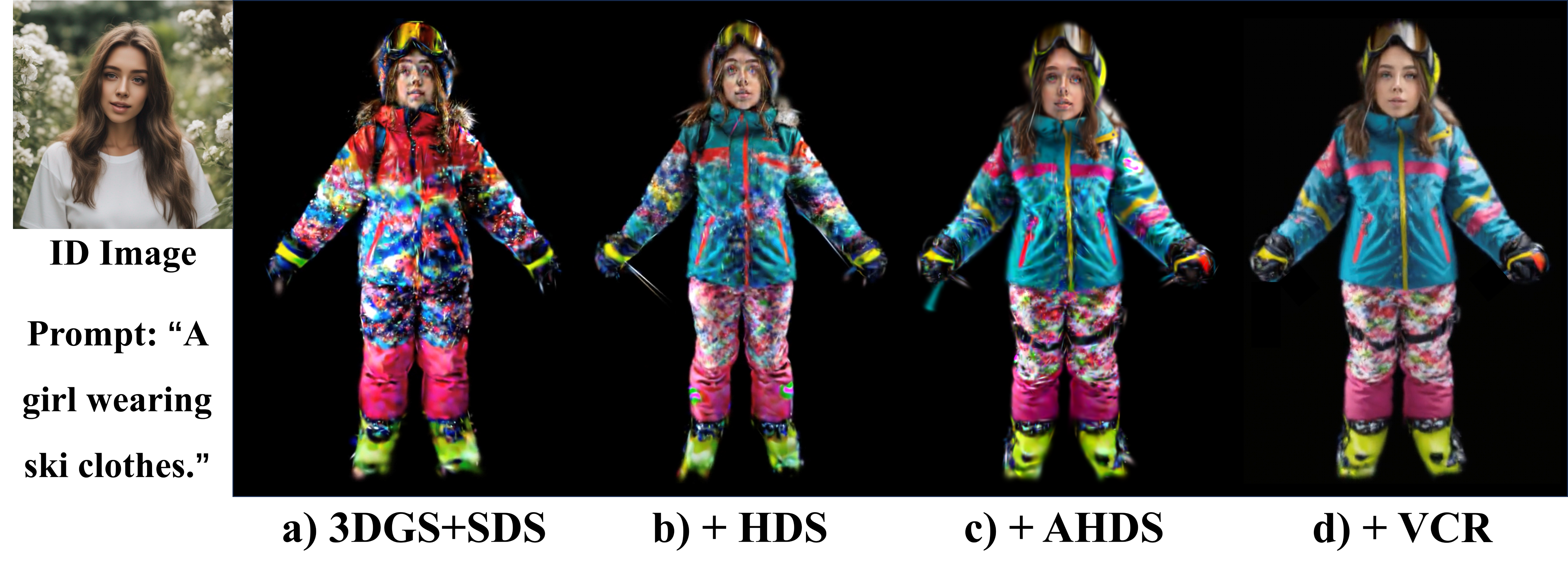}
\caption{Ablation studies on various module designs. We present the generation results of the human frontal view under four ablation settings: (a) baseline; (b) + HDS; (c) + AHDS; (d) + View-consistent Refinement Mechanism. Detailed ablation settings and result analysis are depicted in Sec. \ref{sec:4.3}.}
\label{fig:4}
\end{figure}

\begin{figure}[t]
\centering
\includegraphics[width=1.0\columnwidth]{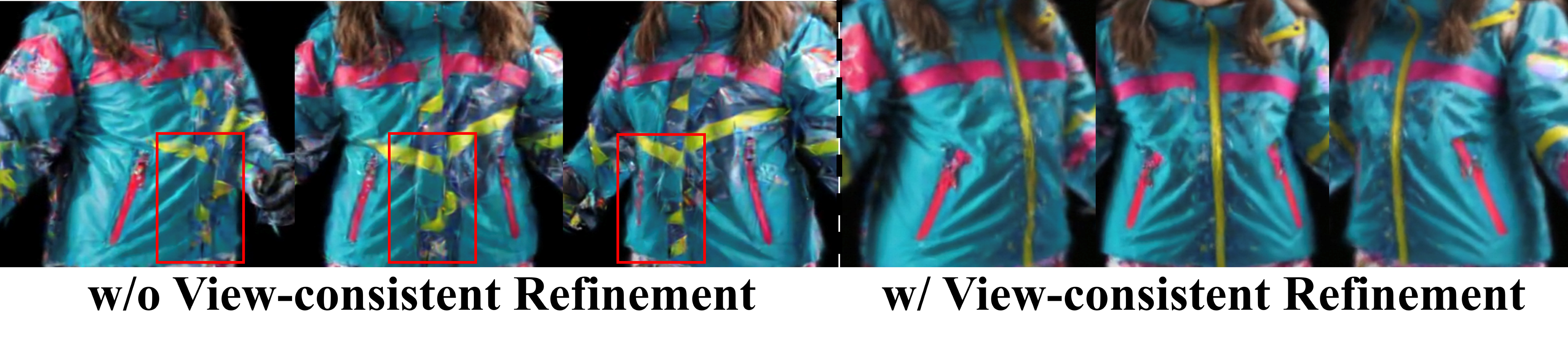}
\caption{Ablation study on our VCR module. When the multi-view images are denoised independently, the results will loss cross-view texture consistency. In contrast, images refined with our VCR module maintain high 3D texture consistency.}
\label{fig:5}
\end{figure}

\subsection{Ablation Studies} \label{sec:4.3}
\noindent \textbf{Ablations for AHDS and Refinement Mechanism.} We evaluate the visual results of \textbf{a)} our baseline approach (3DGS+SDS) and progressively add \textbf{b)} HDS, \textbf{c)} AHDS (HDS with adaptive timestep scheduling), and \textbf{d)} the View-Consistent Refinement (VCR) mechanism. The results in Fig. \ref{fig:4} show that the avatar generated with the baseline forms basic shapes but tends to produce an over-saturated appearance and lacks fine details. Incorporating HDS leads to improvements in identity preservation and clothing details. With the addition of timestep scheduling in AHDS, the training time required for HDS is significantly reduced (with training steps reduced from 3600 to 2400) while achieving enhanced generative quality. The generated results display more intricate textures and detailed facial features. Finally, the refinement mechanism further improves multi-view consistency, achieving high-fidelity, coherent textures across views.

\noindent \textbf{Ablation for View-consistent Refinement.} To assess the importance of multi-view consistency, we perform an ablation on the refinement mechanism. As illustrated in Fig. \ref{fig:5}, when the image of each view undergoes independent denoising process, the result lacks some texture consistency across various views, showing misaligned and blurred details when we change the camera views. In contrast, applying our view-consistent refinement ensures texture alignment across all views, maintaining consistent and realistic details, especially in critical facial and garment features.

\section{Conclusion}
In this paper, we present GaussianIP, a novel two-stage framework for generating realistic 3D human avatars from text and image prompts while preserving identity. We propose AHDS guidance to facilitate the learning of identity information and accelerate the generation process. Additionally, we introduce VCR mechanism to enhance the texture fidelity of 3D avatars without sacrificing 3D consistency. Extensive experiments validate the superiority of GaussianIP in both efficiency and visual details.

\noindent \textbf{Limitations and future work.} Our method faces challenges in rendering highly complex poses or extreme garment textures. Further research is required to generalize the framework to broader applications, such as human-object and human-human interactions. We aim to investigate these aspects in future work, enabling more immersive and richer interactive experiences in VR/AR environments.
\section*{Acknowledgment}
This work is partly supported by the National Key R\&D Program of China (2022ZD0161902), and  the National Natural Science Foundation of China (No. 62202031).  We also give specical thanks to Alibaba Group for their contribution to this paper.
{
    \small
    \bibliographystyle{ieeenat_fullname}
    \bibliography{main}
}


\end{document}